\documentclass[11pt,a4paper]{article}
\usepackage[hyperref]{naaclhlt2019}
\usepackage{times}
\usepackage{latexsym}
\usepackage{url}
\usepackage{graphicx}

\aclfinalcopy 

\title{Experiments in Cuneiform Language Identification}

\author{
  Gustavo Henrique Paetzold\textsuperscript{1}, Marcos Zampieri\textsuperscript{2} \\
  \textsuperscript{1}Universidade Tecnol\'{o}gica Federal do Paran\'{a}, Toledo-PR, Brazil\\
  \textsuperscript{2}University of Wolverhampton, Wolverhampton, United Kingdom
  \\
  {\tt ghpaetzold@utfpr.edu.br} 
}

\date{}

\begin{document}
\maketitle
\begin{abstract}
This paper presents methods to discriminate between languages and dialects written in Cuneiform script, one of the first writing systems in the world. We report the results obtained by the {\em PZ} team in the Cuneiform Language Identification (CLI) shared task organized within the scope of the VarDial Evaluation Campaign 2019. The task included two languages, Sumerian and Akkadian. The latter is divided into six dialects: Old Babylonian, Middle Babylonian peripheral, Standard Babylonian, Neo Babylonian, Late Babylonian, and Neo Assyrian. We approach the task using a meta-classifier trained on various SVM models and we show the effectiveness of the system for this task. Our submission achieved 0.738 F1 score in discriminating between the seven languages and dialects and it was ranked fourth in the competition among eight teams.
\end{abstract}

\section{Introduction}
\label{sec:intro}

As discussed in a recent survey \cite{jauhiainen2018automatic}, discriminating between similar languages, national language varieties, and dialects is an important challenge faced by state-of-the-art language identification systems. The topic has attracted more and more attention from the CL/NLP community in recent years with publications on similar languages of the Iberian peninsula \cite{zubiaga2016tweetlid}, and varieties and dialects of several languages such as Greek \cite{sababa2018classifier} and Romanian \cite{ciobanu2016computational} to name a few.

As evidenced in Section \ref{sec:related}, the focus of most of these studies is the identification of languages and dialects using contemporary data. A few exceptions include the work by \newcite{trieschnigg2012exploration} who applied language identification methods to historical varieties of Dutch and the work by \newcite{CLIarxiv} on languages written in cuneiform script: Sumerian and Akkadian. Cuneiform is an ancient writing system invented by the Sumerians for more than three millennia. 

In this paper we describe computational approaches to language identification on texts written in cuneiform script. For this purpose we use the dataset made available by \newcite{CLIarxiv} to participants of the Cuneiform Language Identification (CLI) shared task organized at VarDial 2019 \cite{vardial2019report}. Our submission, under the team name {\em PZ}, is an adaptation of an n-gram-based meta-classifier system which showed very good performance in previous language identification shared tasks \cite{malmasi-zampieri:2017:VarDial1,malmasi-zampieri:2017:VarDial2}. Furthermore, we compare the performance of the meta-classifier to the submissions to the CLI shared task and, in particular, to a deep learning approach submitted by the team {\em ghpaetzold}. It has been shown in previous language identification studies \cite{medvedeva2017sparse,kroon2018simple} that deep learning approaches do not outperform n-gram-based methods and we were interested in investigating whether this is also true for the languages and dialects included in CLI.

\begin{table*}[!ht]
\center
\begin{tabular}{lcrrr}
\hline
\bf Language or Dialect & \bf Code & \bf Texts & \bf Lines & \bf Signs \\ 
\hline
Late Babylonian & LTB & 671 & 31,893 & ca. 260,000 \\
Middle Babylonian peripheral & MPB & 365 & 11,015 & ca. 95,000 \\
Neo-Assyrian & NE & 3,570 & 65,932 & ca. 490,000 \\
Neo-Babylonian & NEB & 1,212 & 19,414 & ca. 200,000 \\
Old Babylonian & OLB & 527 & 7,605 & ca. 65,000 \\
Standard  Babylonian & STB & 1,661 & 35,633 & ca. 390,000 \\
Sumerian & SUX & 5,000 & 107,345 & ca. 400,000 \\
\hline
Total & & 13,006 & 278,837 & ca. 1,900,000 \\
\hline
\end{tabular}
\caption{Number of texts, lines, and signs in each of the seven languages and dialects in the dataset of \newcite{CLIarxiv}, from which the instances of the CLI datasets were taken.}
\label{tab:CLIdata}
\end{table*}

\section{Related Work}
\label{sec:related}

Since its first edition in 2014, shared tasks on similar language and dialect identification have been organized together with the VarDial workshop co-located with international conferences such as COLING, EACL, and NAACL. The first and most well-attended of these competitions was the Discrminating between Similar Languages (DSL) shared task which has been organized between 2014 and 2017 \cite{dsl2016,zampieri2014vardial,zampieri:2015:LT4VarDial,vardial2017report}. The DSL provided the first benchmark for evaluation of language identification systems developed for similar languages and language varieties using the DSL Corpus Collection (DSLCC) \cite{tan:2014:BUCC}, a multilingual benchmarked dataset compiled for this purpose. In 2017 and 2018, VarDial featured evaluation campaigns with multiple shared tasks not only on language and dialect identification but also on other NLP tasks related to language and dialect variation (e.g. morphosyntactic tagging, and cross-lingual dependency parsing). With the exception of the DSL, the language and dialect identification competitions organized at VarDial focused on groups of dialects from the same language such as Arabic (ADI shared task) and German (GDI shared task).

The focus of the aforementioned language and dialect identification competitions was diatopic variation and thus the data made available in these competitions was synchronic contemporary corpora. In the 2019 edition of the workshop, for the first time, a task including historical languages was organized. The CLI shared task provided participants with a dataset containing languages and dialects written in cuneiform script: Sumerian and Akkadian. Akkadian is divided into six dialects in the dataset: Old Babylonian, Middle Babylonian peripheral, Standard Babylonian, Neo Babylonian, Late Babylonian, and Neo Assyrian \cite{CLIarxiv}. 

The CLI shared task is an innovative initiative that opens new perspectives in the computational processing of languages written in cuneiform script. There have been a number of studies applying computational methods to process these languages (e.g. Sumerian \cite{chiarcos2018towards}), but with the exception of \newcite{CLIarxiv}, to the best of our knowledge, no language identification studies have been published. CLI is the first competition organized on cuneiform script texts in particular and in historical language identification in general.



\section{Methodology and Data}
\label{sec:method}

The dataset used in the CLI shared task is described in detail in \newcite{CLIarxiv}. All of the data included in the dataset was collected from the Open Richly Annotated Cuneiform Corpus (Oracc)\footnote{\url{http://oracc.museum.upenn.edu/}} which contains transliterated texts. \newcite{CLIarxiv} created a tool to transform the texts back to the cuneiform script. The dataset features texts from seven languages and dialects amounting to a little over 13,000 texts. The list of languages and dialects is presented in Table \ref{tab:CLIdata}.

\subsection{System Description}

Our submission to the CLI shared task is a system based on a meta-classifier trained on several SVM models. Meta-classifiers \cite{giraud2004introduction} and ensemble learning methods have proved to deliver competitive performance not only in language identification \cite{malmasi-zampieri:2017:VarDial1,malmasi-zampieri:2017:VarDial2} but also in many other text classification tasks \cite{malmasi2016predicting,sulea2017exploring}. 

The meta-classifier is an adaptation of previous submissions to VarDial shared tasks described in \cite{malmasi-zampieri:2017:VarDial2}. It is essentially a bagging ensemble trained on the outputs of linear SVM classifiers. As features, the system uses the following character n-gram and character skip-gram features:

\begin{itemize}
    \item character $n$-grams of order $1$--$5$;
    \item 1-skip character bigrams and trigrams;
    \item 2-skip character bigrams and trigrams;
    \item 3-skip character bigrams and trigrams.
\end{itemize}

\noindent Each feature class is used to train a single linear SVM classifier using LIBLINEAR \cite{fan2008liblinear}. The outputs of these SVM classifiers on the training data are then used to train the meta-classifier. 


\section{Results}
\label{sec:results}

Table \ref{tab:results-CLI-open} showcases the results obtained by our team ({\em PZ} in bold) and the best submission by each of the eight teams which participating in the CLI shared task. Even though the competition allowed the use of other datasets (open submission), we have used only the dataset provided by the shared task organizers to train our model. 

Our submission was ranked 4\textsuperscript{th} in the shared task, only a few percentage points below the top-3 systems: {\em NRC-CNRC, tearsofjoy}, and {\em Twist Bytes}. The meta-classifier achieved much higher performance at distinguishing between these Mesopotamian languages and dialects than the neural model by {\em ghpaetzold}, which ranked 6\textsuperscript{th} in the competition. We present this neural model in more detail comparing its performance to our meta-classifier in Section \ref{sec:comparison}.

\begin{table}[h]
\center
\begin{tabular}{lc}
\hline
\bf System & \bf F1 (macro) \\ 
\hline
NRC-CNRC & 0.769  \\
tearsofjoy & 0.763  \\
Twist Bytes & 0.743  \\
\bf PZ & \bf 0.738  \\
ghmerti & 0.721 \\
ghpaetzold & 0.556 \\
ekh &	0.550 \\
situx	& 0.128 \\
\hline
\end{tabular}
\caption{Results for the CLI task obtained by the team {\em PZ} (in bold) in comparison to the the best entries of each of the eight teams in the shared task. Results reported in terms of F1 (macro).}
\label{tab:results-CLI-open}
\end{table}

\subsection{Comparison to a Neural Model}
\label{sec:comparison}

We take the opportunity to compare the performance of our system with an entirely different type of model submitted by team {\em ghpaetzold}. This comparison was motivated by the lower performance obtained by the neural models in comparison to traditional machine learning models in previous VarDial shared tasks \cite{vardial2018report}. It was made possible due to the collaboration between the {\em ghpaetzold} team and ours.\footnote{One of the {\em ghpaetzold} team members was also a member of the {\em PZ} team.}

As demonstrated by \newcite{ling2015finding}, compositional recurrent neural networks can offer very reliable performance on a variety of NLP tasks. Previous language identification and dialect studies \cite{medvedeva2017sparse,kroon2018simple,butnaru2019moroco} and the results of the previous shared tasks organized at VarDial \cite{vardial2017report,vardial2018report}, however, showed that deep learning approaches do not outperform more linear n-gram-based methods so we were interested in comparing the performance of a neural model to the meta-classifier for this dataset. 

A compositional network is commonly described as a model that builds numerical representations of words based on the sequence of characters that compose them. They are inherently more time-consuming to train than typical neural models that use traditional word vectors because of the added parameters, but they compensate by being able to handle any conceivable word passed as input with very impressive robustness \cite{paetzold2018utfpr,paetzold}.

The model takes as input a sentence and produces a corresponding label as output. First, the model vectorizes each character of each word in the sentence using a typical character embedding layer. It then passes the sequence of vectors through a set of 2 layers of Gated Recurrent Units (GRUs) and produces a numerical representation for each word as a whole. This set of representations is then passed through another 2-layer set of GRUs to produce a final vector for the sentence as a whole, and then a dense layer is used to produce a softmax distribution over the label set. The model uses 25 dimensions for character embeddings, 30 nodes for each GRU layer and 50\% dropout. A version of each model was saved after each epoch so that the team could choose the one with the lowest error on the development set as their submission. 

\begin{figure*}[!ht]
\begin{minipage}{8cm}
\includegraphics[width=0.95\textwidth]{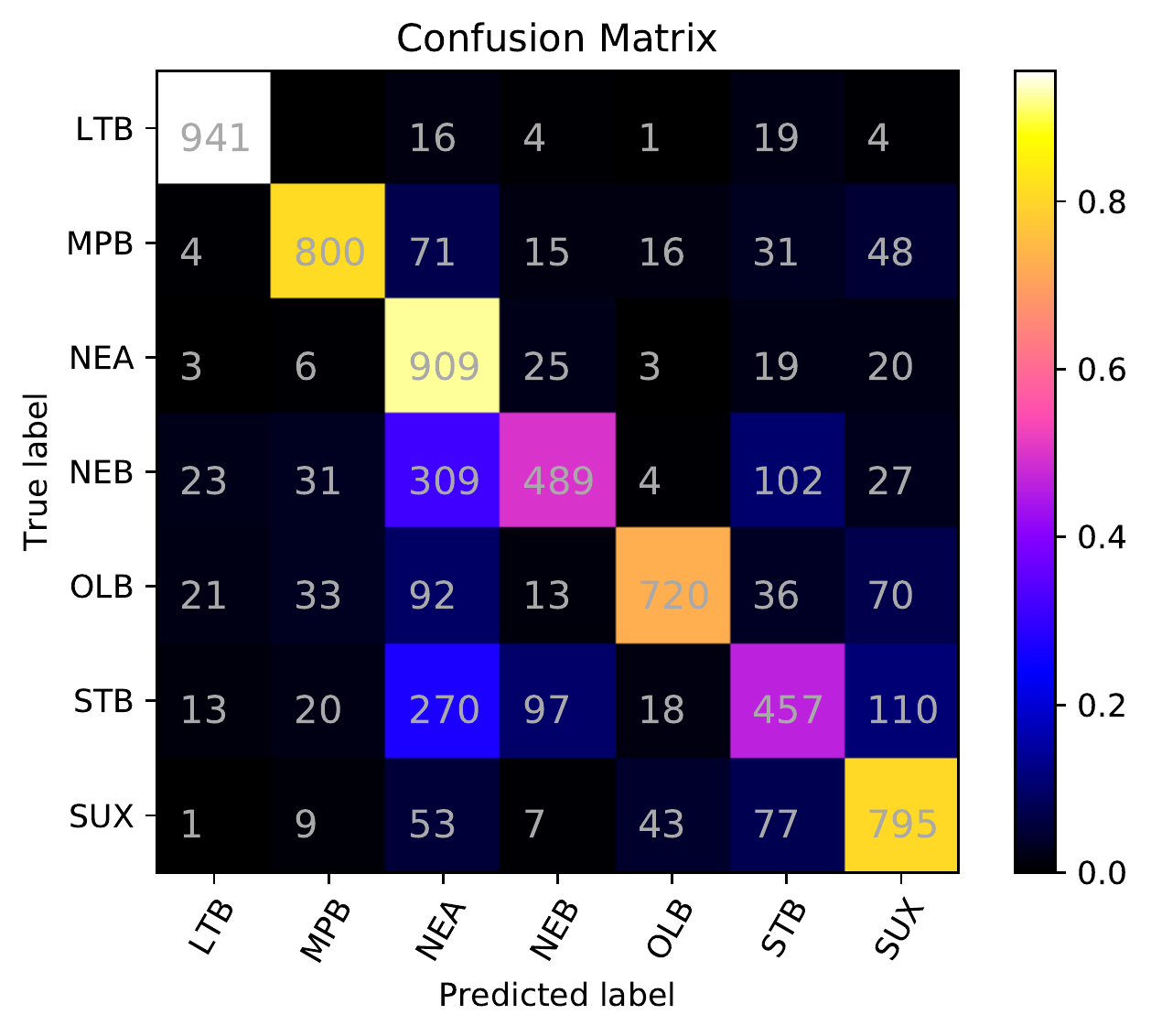}
\caption{Confusion matrix for the meta-classifier.}
\label{fig:1}
\end{minipage}
\begin{minipage}{8cm}
\includegraphics[width=0.95\textwidth]{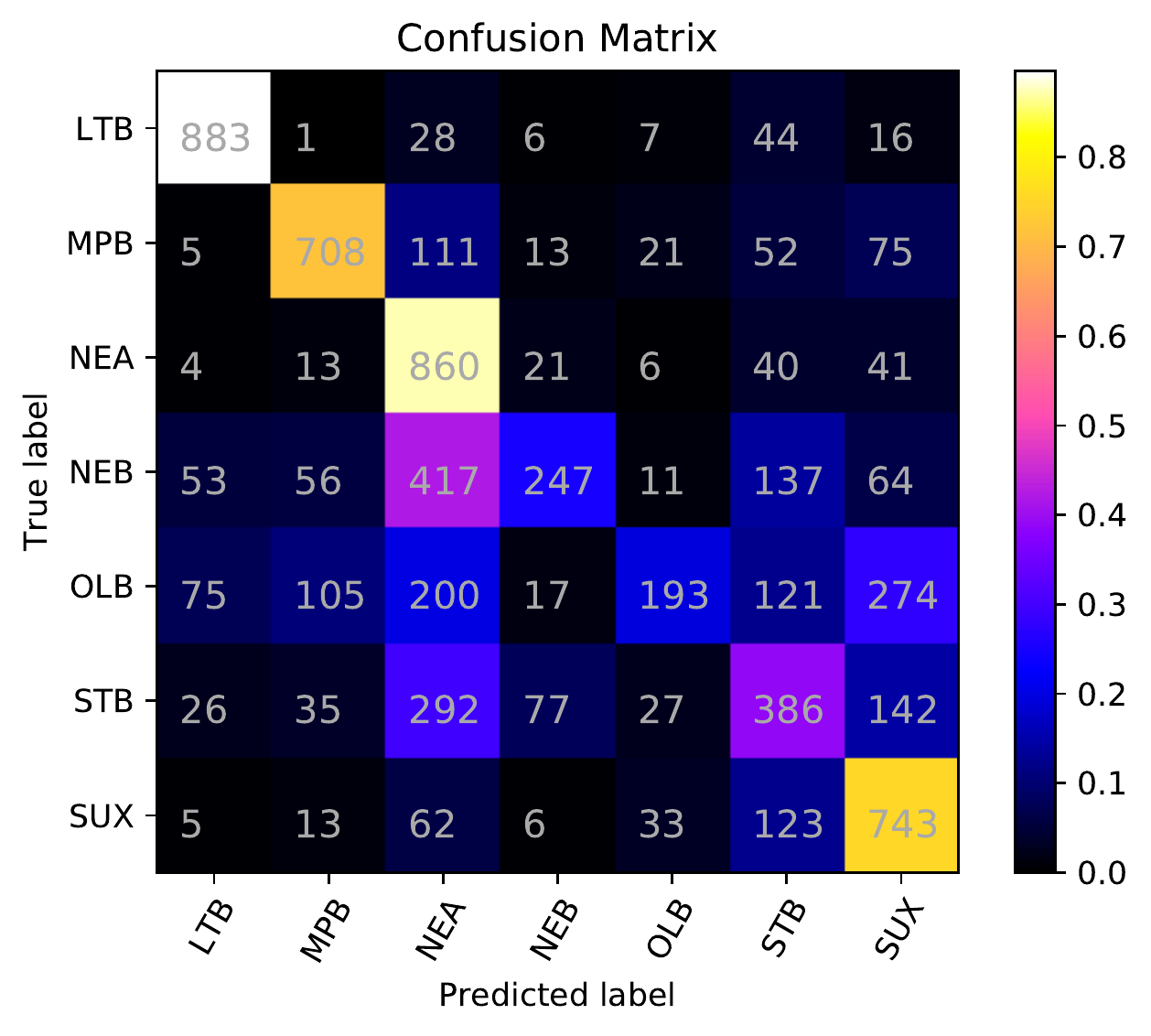}
\caption{Confusion matrix for the neural model.}
\label{fig:2}
\end{minipage}
\end{figure*}

Inspecting the two confusion matrices depicted in Figures \ref{fig:1} and \ref{fig:2}, we found that the neural model did not do very well at differentiating between Standard Babylonian and Neo Assyrian, as well as between Neo Babylonian and Neo Assyrian, leading to many misclassifications. These two language pairs were also the most challenging for the meta-classifier, however, the number of missclassified instances by the meta-classifier was much lower.

\section{Conclusion and Future Work}

In this paper we presented a meta-classifier system submitted by the team {\em PZ} to the Cuneiform Language Identification shared task organized at VarDial 2019. Our submission is an adaptation of a sophisticated meta-classifier which achieved high performance in previous language and dialect identification shared tasks at VarDial \cite{malmasi-zampieri:2017:VarDial2}. The meta-classifier combines the output of multiple SVM classifers trained on character-based features. The meta-classifier ranked 4\textsuperscript{th} in the competition among eight teams only a few percentage points below the top-3 systems in the competition. 

Finally, we compared the performance of the meta-classifier with a compositional RNN model that uses only the text from the instance as input trained on the same dataset. The comparison shows that, while the neural model does offer competitive performance against some of the systems submitted to the shared task, the more elaborate features used by the meta-classifier allows it to much more proficiently distinguish between very similar language pairs, such as Neo Babylonian and Neo Assyrian, leading to a performance gain of 18.2\% F-score and 2 positions in the shared task rankings. The results obtained by the meta-classifier in comparison to the neural model corroborate the findings of previous studies \cite{medvedeva2017sparse} in the last two VarDial evaluation campaigns \cite{vardial2017report,vardial2018report}.

In the future we would like to analyze the results obtained by the highest performing teams in the CLI shared task. The top team achieved the best performance in the competition using a neural-based method. This is, to the best of our knowledge, the first time in which a deep learning approach outperforms traditional machine learning methods in one of the VarDial shared tasks. The great performance obtained by the NRC-CNRC team might be explained by the use of more suitable deep learning methods such as BERT \cite{devlin2018bert}.

\section*{Acknowledgements}

We would like to thank Shervin Malmasi for his valuable suggestions and feedback. We further thank the CLI shared task organizer, Tommi Jauhiainen, for organizing this interesting shared task. 

We gratefully acknowledge the support of NVIDIA Corporation with the donation of the Titan V GPU used for this research.

\bibliography{vardial}

\begin{thebibliography}{27}
\expandafter\ifx\csname natexlab\endcsname\relax\def\natexlab#1{#1}\fi

\bibitem[{Butnaru and Ionescu(2019)}]{butnaru2019moroco}
Andrei Butnaru and Radu~Tudor Ionescu. 2019.
\newblock {MOROCO: The Moldavian and Romanian Dialectal Corpus}.
\newblock \emph{arXiv preprint arXiv:1901.06543}.

\bibitem[{Chiarcos et~al.(2018)Chiarcos, Pag{\'e}-Perron, Khait, Schenk, and
  Reckling}]{chiarcos2018towards}
Christian Chiarcos, {\'E}milie Pag{\'e}-Perron, Ilya Khait, Niko Schenk, and
  Lucas Reckling. 2018.
\newblock {Towards a Linked Open Data Edition of Sumerian Corpora}.
\newblock In \emph{Proceedings of LREC}.

\bibitem[{Ciobanu and Dinu(2016)}]{ciobanu2016computational}
Alina~Maria Ciobanu and Liviu~P Dinu. 2016.
\newblock A computational perspective on the romanian dialects.
\newblock In \emph{Proceedings of LREC}.

\bibitem[{Devlin et~al.(2018)Devlin, Chang, Lee, and
  Toutanova}]{devlin2018bert}
Jacob Devlin, Ming-Wei Chang, Kenton Lee, and Kristina Toutanova. 2018.
\newblock {BERT: Pre-training of Deep Bidirectional Transformers for Language
  Understanding}.
\newblock \emph{arXiv preprint arXiv:1810.04805}.

\bibitem[{Fan et~al.(2008)Fan, Chang, Hsieh, Wang, and Lin}]{fan2008liblinear}
Rong-En Fan, Kai-Wei Chang, Cho-Jui Hsieh, Xiang-Rui Wang, and Chih-Jen Lin.
  2008.
\newblock {LIBLINEAR: A Library for Large Linear Classification}.
\newblock \emph{Journal of Machine Learning Research}, 9(Aug):1871--1874.

\bibitem[{Giraud-Carrier et~al.(2004)Giraud-Carrier, Vilalta, and
  Brazdil}]{giraud2004introduction}
Christophe Giraud-Carrier, Ricardo Vilalta, and Pavel Brazdil. 2004.
\newblock {Introduction to the Special Issue on Meta-learning}.
\newblock \emph{Machine learning}, 54(3):187--193.

\bibitem[{Jauhiainen et~al.(2019)Jauhiainen, Jauhiainen, Alstola, and
  Lind{\'e}n}]{CLIarxiv}
Tommi Jauhiainen, Heidi Jauhiainen, Tero Alstola, and Krister Lind{\'e}n. 2019.
\newblock {Language and Dialect Identification of Cuneiform Texts}.
\newblock In \emph{Proceedings of VarDial}.

\bibitem[{Jauhiainen et~al.(2018)Jauhiainen, Lui, Zampieri, Baldwin, and
  Lind{\'e}n}]{jauhiainen2018automatic}
Tommi Jauhiainen, Marco Lui, Marcos Zampieri, Timothy Baldwin, and Krister
  Lind{\'e}n. 2018.
\newblock Automatic language identification in texts: A survey.
\newblock \emph{arXiv preprint arXiv:1804.08186}.

\bibitem[{Kroon et~al.(2018)Kroon, Medvedeva, and Plank}]{kroon2018simple}
Martin Kroon, Masha Medvedeva, and Barbara Plank. 2018.
\newblock {When Simple N-gram Models Outperform Syntactic Approaches:
  Discriminating between Dutch and Flemish}.
\newblock In \emph{Proceedings of VarDial}.

\bibitem[{Ling et~al.(2015)Ling, Lu{\'\i}s, Marujo, Astudillo, Amir, Dyer,
  Black, and Trancoso}]{ling2015finding}
Wang Ling, Tiago Lu{\'\i}s, Lu{\'\i}s Marujo, Ram{\'o}n~Fernandez Astudillo,
  Silvio Amir, Chris Dyer, Alan~W Black, and Isabel Trancoso. 2015.
\newblock Finding function in form: Compositional character models for open
  vocabulary word representation.
\newblock \emph{arXiv preprint arXiv:1508.02096}.

\bibitem[{Malmasi and
  Zampieri(2017{\natexlab{a}})}]{malmasi-zampieri:2017:VarDial2}
Shervin Malmasi and Marcos Zampieri. 2017{\natexlab{a}}.
\newblock {Arabic Dialect Identification Using iVectors and ASR Transcripts}.
\newblock In \emph{Proceedings of VarDial}.

\bibitem[{Malmasi and
  Zampieri(2017{\natexlab{b}})}]{malmasi-zampieri:2017:VarDial1}
Shervin Malmasi and Marcos Zampieri. 2017{\natexlab{b}}.
\newblock {German Dialect Identification in Interview Transcriptions}.
\newblock In \emph{Proceedings of VarDial}.

\bibitem[{Malmasi et~al.(2016{\natexlab{a}})Malmasi, Zampieri, and
  Dras}]{malmasi2016predicting}
Shervin Malmasi, Marcos Zampieri, and Mark Dras. 2016{\natexlab{a}}.
\newblock {Predicting Post Severity in Mental Health Forums}.
\newblock In \emph{Proceedings of CLPsych}.

\bibitem[{Malmasi et~al.(2016{\natexlab{b}})Malmasi, Zampieri,
  Ljube\v{s}i\'{c}, Nakov, Ali, and Tiedemann}]{dsl2016}
Shervin Malmasi, Marcos Zampieri, Nikola Ljube\v{s}i\'{c}, Preslav Nakov, Ahmed
  Ali, and J\"{o}rg Tiedemann. 2016{\natexlab{b}}.
\newblock {Discriminating between Similar Languages and Arabic Dialect
  Identification: A Report on the Third DSL Shared Task}.
\newblock In \emph{Proceedings of VarDial}.

\bibitem[{Medvedeva et~al.(2017)Medvedeva, Kroon, and
  Plank}]{medvedeva2017sparse}
Maria Medvedeva, Martin Kroon, and Barbara Plank. 2017.
\newblock {When Sparse Traditional Models Outperform Dense Neural Networks: The
  Curious Case of Discriminating between Similar Languages}.
\newblock In \emph{Proceedings of VarDial}.

\bibitem[{Paetzold(2018)}]{paetzold2018utfpr}
Gustavo Paetzold. 2018.
\newblock {UTFPR at IEST 2018: Exploring Character-to-Word Composition for
  Emotion Analysis}.
\newblock In \emph{Proceedings of WASSA}.

\bibitem[{Paetzold(2019)}]{paetzold}
Gustavo Paetzold. 2019.
\newblock {UTFPR at SemEval-2019 Task 6: Relying on Compositionality to Find
  Offense}.
\newblock In \emph{Proceedings of SemEval}.

\bibitem[{Sababa and Stassopoulou(2018)}]{sababa2018classifier}
Hanna Sababa and Athena Stassopoulou. 2018.
\newblock {A Classifier to Distinguish Between Cypriot Greek and Standard
  Modern Greek}.
\newblock In \emph{Proceedings of SNAMS}.

\bibitem[{Sulea et~al.(2017)Sulea, Zampieri, Malmasi, Vela, Dinu, and van
  Genabith}]{sulea2017exploring}
Octavia-Maria Sulea, Marcos Zampieri, Shervin Malmasi, Mihaela Vela, Liviu~P
  Dinu, and Josef van Genabith. 2017.
\newblock {Exploring the use of Text Classification in the Legal Domain}.
\newblock \emph{Proceedings of ASAIL}.

\bibitem[{Tan et~al.(2014)Tan, Zampieri, Ljube\v{s}i\'{c}, and
  Tiedemann}]{tan:2014:BUCC}
Liling Tan, Marcos Zampieri, Nikola Ljube\v{s}i\'{c}, and J{\"o}rg Tiedemann.
  2014.
\newblock {Merging Comparable Data Sources for the Discrimination of Similar
  Languages: The DSL Corpus Collection}.
\newblock In \emph{Proceedings BUCC}.

\bibitem[{Trieschnigg et~al.(2012)Trieschnigg, Hiemstra, Theune, Jong, and
  Meder}]{trieschnigg2012exploration}
Dolf Trieschnigg, Djoerd Hiemstra, Mari{\"e}t Theune, Franciska Jong, and Theo
  Meder. 2012.
\newblock {An Exploration of Language Identification Techniques in the Dutch
  Folktale Database}.
\newblock In \emph{Proceedings of the Workshop on Adaptation of Language
  Resources and Tools for Processing Cultural Heritage}.

\bibitem[{Zampieri et~al.(2017)Zampieri, Malmasi, Ljube\v{s}i\'{c}, Nakov, Ali,
  Tiedemann, Scherrer, and Aepli}]{vardial2017report}
Marcos Zampieri, Shervin Malmasi, Nikola Ljube\v{s}i\'{c}, Preslav Nakov, Ahmed
  Ali, J\"{o}rg Tiedemann, Yves Scherrer, and No\"{e}mi Aepli. 2017.
\newblock {Findings of the VarDial Evaluation Campaign 2017}.
\newblock In \emph{Proceedings of VarDial}.

\bibitem[{Zampieri et~al.(2018)Zampieri, Malmasi, Nakov, Ali, Shuon, Glass,
  Scherrer, Samard\v{z}i{\'c}, Ljube\v{s}i\'{c}, Tiedemann, {van der Lee},
  Grondelaers, Oostdijk, {van den Bosch}, Kumar, Lahiri, and
  Jain}]{vardial2018report}
Marcos Zampieri, Shervin Malmasi, Preslav Nakov, Ahmed Ali, Suwon Shuon, James
  Glass, Yves Scherrer, Tanja Samard\v{z}i{\'c}, Nikola Ljube\v{s}i\'{c},
  J\"{o}rg Tiedemann, Chris {van der Lee}, Stefan Grondelaers, Nelleke
  Oostdijk, Antal {van den Bosch}, Ritesh Kumar, Bornini Lahiri, and Mayank
  Jain. 2018.
\newblock {Language Identification and Morphosyntactic Tagging: The Second
  VarDial Evaluation Campaign}.
\newblock In \emph{Proceedings of VarDial}.

\bibitem[{Zampieri et~al.(2019)Zampieri, Malmasi, Scherrer,
  Samard{\v{z}}i{\'{c}}, Tyers, Silfverberg, Klyueva, Pan, Huang, Ionescu,
  Butnaru, and Jauhiainen}]{vardial2019report}
Marcos Zampieri, Shervin Malmasi, Yves Scherrer, Tanja Samard{\v{z}}i{\'{c}},
  Francis Tyers, Miikka Silfverberg, Natalia Klyueva, Tung-Le Pan, Chu-Ren
  Huang, Radu~Tudor Ionescu, Andrei Butnaru, and Tommi Jauhiainen. 2019.
\newblock {A Report on the Third VarDial Evaluation Campaign}.
\newblock In \emph{Proceedings of VarDial}.

\bibitem[{Zampieri et~al.(2014)Zampieri, Tan, Ljube\v{s}i\'{c}, and
  Tiedemann}]{zampieri2014vardial}
Marcos Zampieri, Liling Tan, Nikola Ljube\v{s}i\'{c}, and J\"{o}rg Tiedemann.
  2014.
\newblock A report on the {DSL} shared task 2014.
\newblock In \emph{Proceedings of VarDial}.

\bibitem[{Zampieri et~al.(2015)Zampieri, Tan, Ljube\v{s}i\'{c}, Tiedemann, and
  Nakov}]{zampieri:2015:LT4VarDial}
Marcos Zampieri, Liling Tan, Nikola Ljube\v{s}i\'{c}, J\"{o}rg Tiedemann, and
  Preslav Nakov. 2015.
\newblock Overview of the dsl shared task 2015.
\newblock In \emph{Proceedings of LT4VarDial}.

\bibitem[{Zubiaga et~al.(2016)Zubiaga, San~Vicente, Gamallo, Pichel, Alegria,
  Aranberri, Ezeiza, and Fresno}]{zubiaga2016tweetlid}
Arkaitz Zubiaga, Inaki San~Vicente, Pablo Gamallo, Jos{\'e}~Ramom Pichel, Inaki
  Alegria, Nora Aranberri, Aitzol Ezeiza, and V{\'\i}ctor Fresno. 2016.
\newblock {Tweetlid: A Benchmark for Tweet Language Identification}.
\newblock \emph{Language Resources and Evaluation}, 50(4):729--766.

\end{thebibliography}
\bibliographystyle{acl_natbib}

\end{document}